\documentclass[runningheads]{llncs}
\usepackage{graphicx}
\usepackage{amsmath,amssymb} % define this before the line numbering.
\usepackage{color}
\usepackage[width=122mm,left=12mm,paperwidth=146mm,height=193mm,top=12mm,paperheight=217mm]{geometry}

\usepackage{multirow}
\usepackage{booktabs}
\usepackage{caption}
\usepackage{subcaption}

\usepackage[pagebackref=true,colorlinks,bookmarks=false]{hyperref}

\def\eg{\emph{e.g. }} 
\def\ie{\emph{i.e. }}

\def\etal{\emph{et al. }}

\begin{document}
% \linenumbers
\pagestyle{headings}
\mainmatter

\title{Fashion Landmark Detection in the Wild} % Replace with your title

\titlerunning{Fashion Landmark Detection}

\authorrunning{Z. Liu \etal}

\author{Ziwei Liu$^{1\star}$, Sijie Yan$^1$\thanks{The first two authors contribute equally and share first authorship.}, Ping Luo$^{2,1}$, Xiaogang Wang$^{1,2}$, Xiaoou Tang$^{1,2}$}

%Please write out author names in full in the paper, i.e. full given and family names.
%If any authors have names that can be parsed into FirstName LastName in multiple ways, please include the correct parsing, in a comment to the volume editors:
%\index{Lastnames, Firstnames}
%(Do not uncomment it, because you may introduce extra index items if you do that...)

\institute{Dept. of Information Engineering, The Chinese University of Hong Kong\and
Shenzhen Key Lab of Comp. Vis. \& Pat. Rec., Shenzhen Institutes of Advanced Technology, CAS, China\\
\email{\{lz013,siyan,pluo,xtang\}@ie.cuhk.edu.hk, xgwang@ee.cuhk.edu.hk}
}

\maketitle

\begin{abstract}

Visual fashion analysis has attracted many attentions in the recent years. Previous work represented clothing regions by either bounding boxes or human joints. This work presents fashion landmark detection or fashion alignment, which is to predict the positions of functional key points defined on the fashion items, such as the corners of neckline, hemline, and cuff. To encourage future studies, we introduce a fashion landmark dataset\footnote[1]{The dataset is available at \url{http://mmlab.ie.cuhk.edu.hk/projects/DeepFashion/LandmarkDetection.html}} with over 120K images, where each image is labeled with eight landmarks. With this dataset, we study fashion alignment by cascading multiple convolutional neural networks in three stages. These stages gradually improve the accuracies of landmark predictions. Extensive experiments demonstrate the effectiveness of the proposed method, as well as its generalization ability to pose estimation. Fashion landmark is also compared to clothing bounding boxes and human joints in two applications, fashion attribute prediction and clothes retrieval, showing that fashion landmark is a more discriminative representation to understand fashion images.


\end{abstract}

\section{Introduction}

Visual fashion analysis has drawn lots of attentions recently, due to its wide spectrum of applications such as clothes recognition \cite{huang2015cross,kiapour2015where,liu2016deepfashion}, retrieval \cite{liu2012street,di2013style,liu2016deepfashion}, and recommendation \cite{kiapour2014hipster,SimoCVPR15}.
It is a challenging task because of the large variations presented in the clothing items, such as the changes of poses, scales, and appearances.
To reduce these variations, existing works tackled the problem by looking for informative regions, \ie detecting the clothes bounding boxes \cite{huang2015cross,kiapour2015where} or the human joints \cite{chen2012describing,bossard2013apparel}.
We go beyond the above by studying a more discriminative representation, fashion landmark, which is the key-point located at the functional region of clothes, for example the neckline and the cuff.

\begin{figure}[t]
  \centering
  \includegraphics[width=0.9\textwidth]{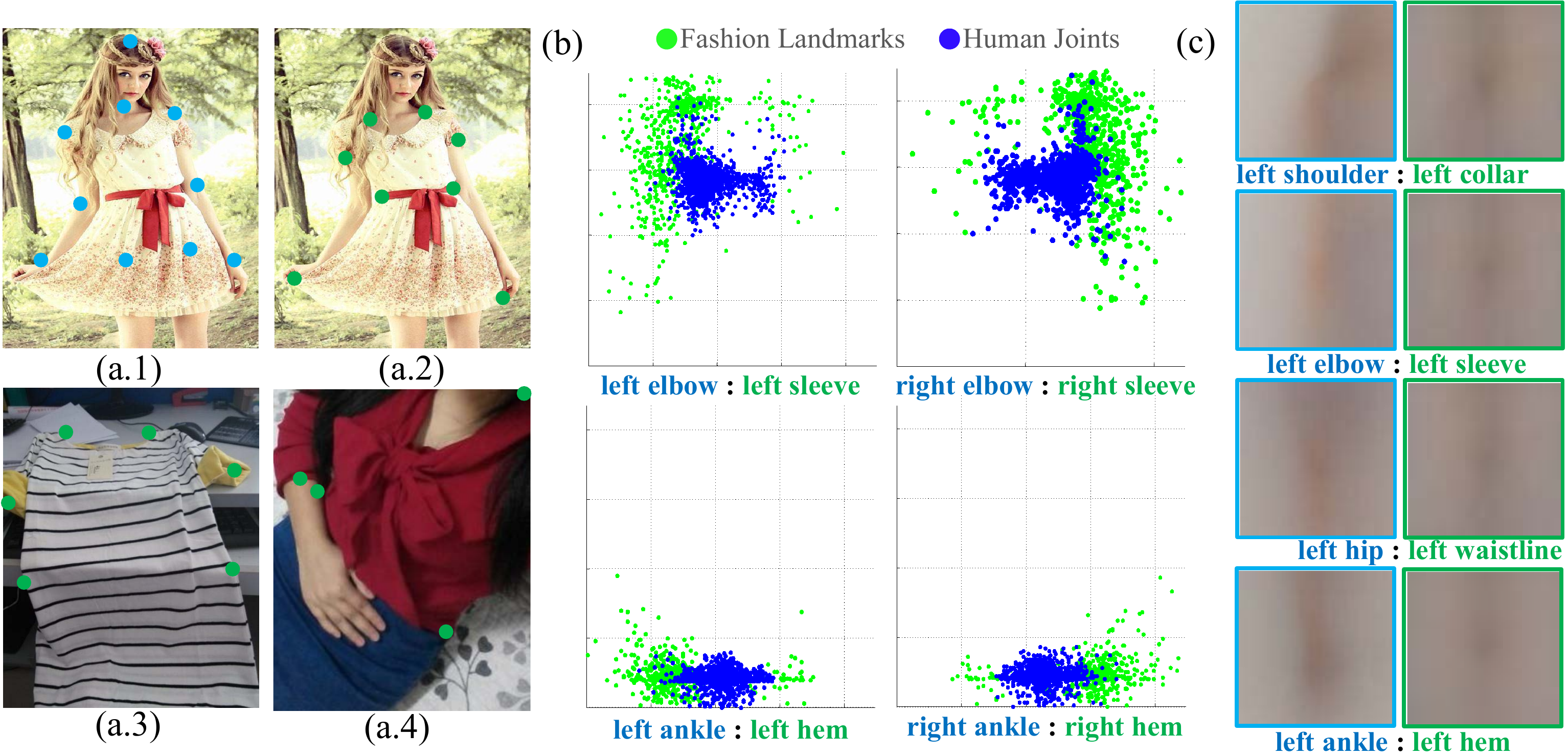}
  \caption{\footnotesize Comparisons of fashion landmarks and human joints: (a.1) sample annotations for human joints, (a.2) sample annotations for fashion landmarks (a.3-4) typical deformation and scale variations present in clothing items, (b) spatial variances of human joints (in blue) and fashion landmarks (in green) and (c) appearance variances of human joints (in blue) and fashion landmarks (in green).}
  \label{fig:intro}
\end{figure}

This work addresses fashion landmark detection or fashion alignment in the wild.
Different from human pose estimation, which detects human joints such as neck and elbows as shown in Fig.\ref{fig:intro} (a.1), fashion alignment localizes fashion landmarks as shown in (a.2).
These landmarks facilitate fashion analysis in the sense that they not only implicitly capture bounding boxes of clothes, but also indicate their functional regions, which can better distinguish design/pattern/category of the clothes.
Therefore, features extracted from these landmarks are more discriminative than those extracted from human joints. For example, when search for a dress with `V-neck and fringed-hem', it is more desirable to extract features from collar and hemline.

To fully benchmark the task of fashion landmark detection, we select a large subset of images from the DeepFashion database \cite{liu2016deepfashion} to constitute a fashion landmark dataset (FLD).
These images have large pose and scale variations.
With FLD, we show that fashion landmark detection in clothes images is a more challenging task than human joint detection in three aspects.
First, clothes undergo \emph{non-rigid deformations} or \emph{scale variations} as shown in Fig.\ref{fig:intro} (a.3-4), while rigid deformations are usually presented in human joints.
Second, fashion landmarks exhibit much larger \emph{spatial variances} than human joints, as illustrated in Fig.\ref{fig:intro} (b), where
we plot the positions of the landmarks and the relative human joints in the test set of the FLD dataset. For instance, the positions of `left sleeve' are more diverse than those of the `left elbow' in both the vertical and horizontal directions.
Third, the local regions of fashion landmarks also have larger \emph{appearance variances} than those of human joints.
As shown in Fig.\ref{fig:intro} (c), we average the patches centered at the fashion landmarks and human joints respectively, resulting in several visual comparisons.
The patterns of the mean patches of human joints are still recognizable, but those of the mean patches of fashion landmarks are not.

To tackle the above challenges, we propose a deep fashion alignment (DFA) framework, which cascades three deep convolutional networks (CNNs) for landmark estimation.
It has three appealing properties.
First, to ensure the CNNs have high discriminative powers, unlike existing works \cite{toshev2014deeppose,chen2014articulated,tompson2014joint,carreira2015human,pfister2015flowing} that only estimated the landmarks' positions, we train the cascaded CNNs to predict both the landmarks' positions and the pseudo-labels, which encode the similarities between training samples to boost the estimation accuracy.
In each stage of the network cascade, the scheme of pseudo-label is carefully designed to reduce different variations presented in the fashion images.
Second, instead of training multiple networks for each body part as previous work did \cite{toshev2014deeppose,fan2015combining}, the DFA framework trains CNNs using the full image as input to significantly reduce computations.
Third, in the DFA framework, an auto-routing strategy is introduced to partition the challenging and easy samples, such that different samples can be handled by different branches of CNNs.

Extensive experiments demonstrate the effectiveness of the proposed method, as well as its generalization ability to pose estimation.
Fashion landmark is also compared to clothing bounding boxes and human joints in two applications, fashion attribute prediction and clothes retrieval, showing that fashion landmark is a more discriminative representation to understand fashion images.

\subsection{Related Work}

\textbf{Visual Fashion Understanding}~
Visual fashion understanding has been a long-pursuing topic due to the many human-centric applications it enables.
Recent advances include predicting semantic attributes \cite{wang2011clothes,chen2012describing,bossard2013apparel,chen2015deep,liu2016deepfashion}, clothes recognition and retrieval \cite{liu2012street,yamaguchi2013paper,kalantidis2013getting,fu2013efficient,kiapour2015where,liu2016deepfashion} and fashion trends discovery \cite{kiapour2014hipster,yamaguchi2014chic,SimoCVPR15}.
To better capture discriminative information in fashion items, previous works have explored the usage of full image \cite{kiapour2015where}, general object proposals \cite{kiapour2015where}, bounding boxes \cite{bossard2013apparel,chen2015deep} and even masks \cite{yamaguchi2012parsing,yamaguchi2013paper,yang2014clothing,liang2015human}.
However, these representations either lack sufficient discriminative ability or are too expensive to obtain.
To overcome these drawbacks, we introduce the problem of clothes alignment in this work, which is a necessary step toward robust fashion recognition.

\textbf{Human Pose Estimation}~
We further convert the problem of clothes alignment into fashion landmark estimation.
Though there are no prior work in fashion landmarks, approaches from similar fields (\eg human pose estimation \cite{ferrari2008progressive,yang2011articulated,dantone2013human,sapp2013modec,toshev2014deeppose,chen2014articulated}) serve as good candidates to explore.
Recently, deep learning \cite{toshev2014deeppose,chen2014articulated,tompson2014joint,belagiannis2015robust} has shown great advantages in locating human joints and there are generally two directions here.
The first direction \cite{toshev2014deeppose,carreira2015human,ramakrishna2014pose} utilizes the power of cascading for iterative error correction.
DeepPose \cite{toshev2014deeppose} employs devide-and-conquer strategy and designs a cascaded deep regression framework on part level
while Iterative Error Feedback \cite{carreira2015human} emphasizes more on the target scheduling in each stage.
The second direction \cite{chen2014articulated,tompson2014joint,dantone2014body,fu2015beyond}, on the other hand, focuses on the explicit modeling of landmark relationships using graphical models.
Chen \etal \cite{chen2014articulated} proposed the combination of CNN and structural SVM \cite{tsochantaridis2004support} to model the tree-like relationships among landmarks
while Thompson \etal \cite{tompson2014joint} plugged Markov Random Field (MRF) \cite{jordan1999introduction} into CNN for joint training.
Here, our framework attempts to absorb the advantages of both directions.
The cascading and auto-routing mechanisms enable both stage-wise and branch-wise variation reductions
while pseudo-labels encode multi-level sample relationships which depict typical global and local landmark configurations.

\begin{figure}[t]
  \centering
  \includegraphics[width=0.9\textwidth]{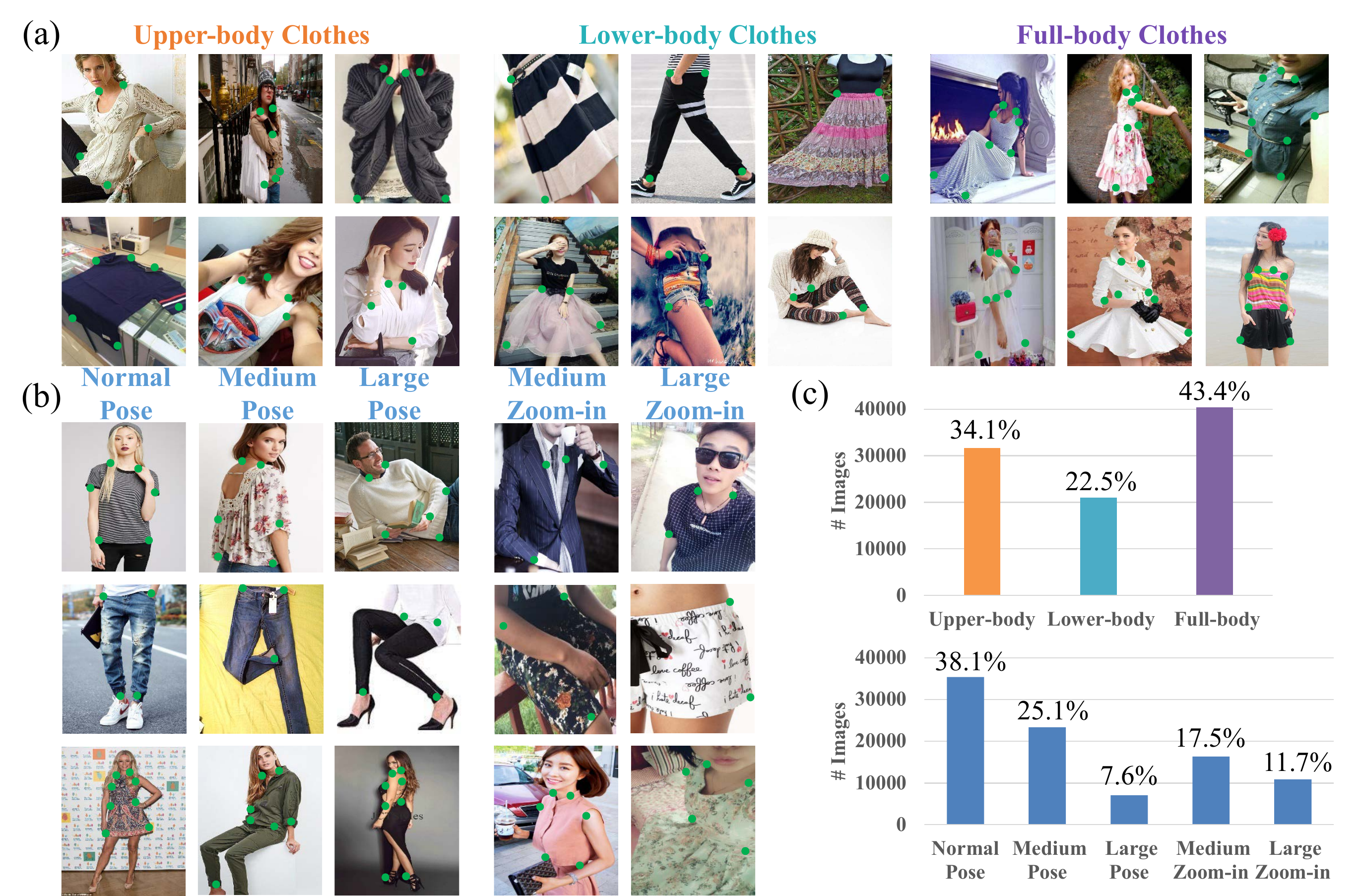}
  \caption{\footnotesize Illustration of the Fashion Landmark Dataset (FLD): (a) sample images and annotations for different types of clothing items, including upper/lower/full-body clothes, (b) sample images and annotations for different subsets, including normal/medium/large poses and medium/large scales, (c) quantitative data distributions.}
  \label{fig:dataset}
\end{figure}

\section{Fashion Landmark Dataset (FLD)}

To benchmark fashion landmark detection, we select a subset of images with large pose and scale variations from DeepFashion database \cite{liu2016deepfashion}, to constitute FLD.
We label and refine landmark annotations in FLD to make sure each image is correctly labeled with $8$ fashion landmarks along with their visibility\footnote{Three states of visibility are defined for each landmark, including visible (located inside of the image and visible), invisible (inside of the image but occluded), and truncated/cut-off (outside of the image).}.
Overall, FLD contains more than $120K$ images. Sample images and annotations are shown in Fig.\ref{fig:dataset} (a).
To characterize the properties of FLD, we divide the images into five subsets according to the positions and visibility of their ground truth landmarks, including the subsets of normal/medium/large poses and medium/large zoom-ins.
The `normal' subset represents images with frontal pose and no cut-off landmarks.
The subsets of `medium' and `large' poses contain images with side or back views,
while the subsets of `medium' and `large' zoom-ins contain clothing items with more than one or three cut-off landmarks, respectively.
Sample images of the five subsets are illustrated in Fig.\ref{fig:dataset} (b) and their statistics are demonstrated in Fig.\ref{fig:dataset} (c), which
shows that FLD contains substantial percentages of images with large poses/scales.

\section{Our Approach}

\begin{figure}[t]
  \centering
  \includegraphics[width=0.9\textwidth]{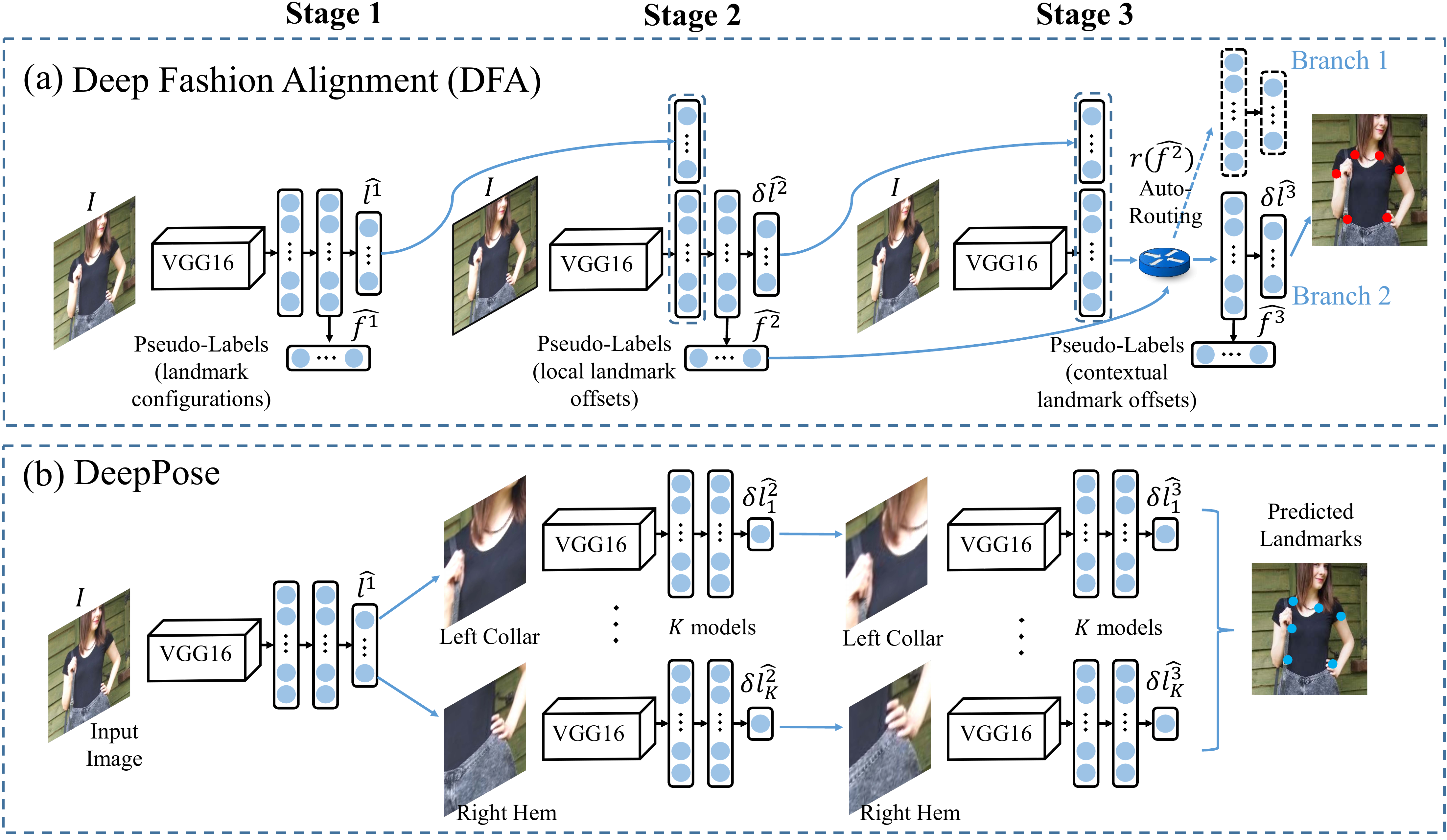}
  \caption{\footnotesize Pipeline of (a) Deep Fashion Alignment (DFA) and (b) DeepPose \cite{toshev2014deeppose}. DFA leverages pseudo-labels and auto-routing mechanism to reduce the large variations presented in fashion images with much less computational costs.}
  \label{fig:pipeline}
\end{figure}

Fashion landmark exhibits large variations in both spatial and appearance domain (see Fig.\ref{fig:intro} (b)(c)).
Fig.\ref{fig:dataset} (c) further shows that more than $30\%$ images have large pose and zoom-in variations.
To account for these challenges, we propose a deep fashion alignment (DFA) framework as shown in Fig.\ref{fig:pipeline} (a), which consists of three stages, where each stage subsequently refines previous predictions.
Unlike the existing representative model for human joints prediction, such as DeepPose \cite{toshev2014deeppose} as shown in (b), which trained multiple networks for all localized parts in each stage, the proposed DFA framework that functions on full image is able to achieve superior performance with much lower computations.

\textbf{Framework Overview}~
As shown in Fig.\ref{fig:pipeline} (a), DFA has three stages. In each stage, VGG-16 \cite{simonyan2014very} is employed as network architecture.
In the first stage, DFA takes the raw image $I$ as input and predicts rough landmark positions, denoted as $\hat{l^1}$, as well as pseudo-labels, denoted as $\hat{f^1}$, which represent landmark configurations such as clothing categories and poses.
In the second stage, both the input image $I$ and the predictions of stage-1, $\hat{l^1}$, are fed in. The whole network is required to predict landmark offsets, signified as $\hat{\delta l^2}$, and pseudo-labels $\hat{f^2}$ that represent local landmark offsets. The landmark prediction of stage-2 is computed as $\hat{l^2} = \hat{l^1} + \hat{\delta l^2}$.
The third stage has two CNNs as two branches, which have identical input and output. Similar to the second stage, each CNN employs image $I$ as input and learns to estimate landmark offsets $\hat{\delta l^3}$ and pseudo-labels $\hat{f^3}$, which contains information about contextual landmark offsets.
In stage-3, each image is passed through one of these two branches. The selection of branch is determined by the predicted pseudo-labels $\hat{f^2}$ in stage-2.
The final prediction is computed as $\hat{l^3} = \hat{l^2} + \hat{\delta l^3}$.

\begin{figure}[t]
  \centering
  \includegraphics[width=0.9\textwidth]{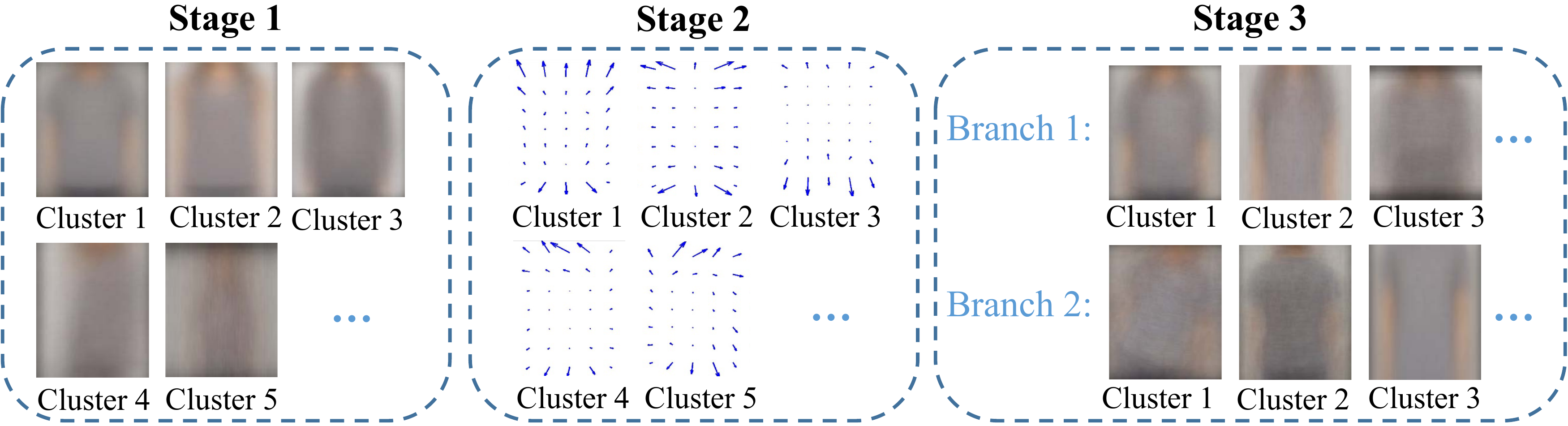}
  \caption{\footnotesize Visualization of pseudo-labels obtained for each stage. Pseudo-labels in stage-1 indicate clothing categories and poses; pseudo-labels in stage-2 represent typical error patterns; pseudo-labels in stage-3 partition within `easy' and `hard' samples.}
  \label{fig:methodology}
\end{figure}

\textbf{Network Cascade}~
Cascade \cite{viola2001rapid} has been proven an effective technique for reducing variations sequentially in pose estimation \cite{toshev2014deeppose}.
Here, we build the DFA system by cascading three CNNs.
The first CNN directly regresses landmark positions and visibility from input images, aided by pseudo-labels in the space of landmark configuration.
These pseudo-labels are achieved by clustering absolute landmark positions, indicating different clothing categories and poses, as shown in Fig.\ref{fig:methodology} stage 1. For example, cluster 1 and 4 represents `short T-shirt in front view' and `long coat in side view', respectively.
The second CNN takes both the input image and the predictions from stage-1 as input, and estimates the offsets that should be made to correct the predictions of the first stage.
In this case, we are learning in the space of landmark offset.
Thus, the pseudo-labels generated here represent typical error patterns and their magnitudes, as shown in Fig.\ref{fig:methodology} stage 2. For instance, cluster 1 represents the corrections should be made along upside and downside, while cluster 2 suggests left/right-direction corrections.
In stage-3, we partition data into two subsets, according to the error patterns predicted in the second stage as shown in Fig.\ref{fig:methodology} stage 3, where branch one deals with `easy' samples such as frontal T-shirt with different sleeve length, while branch two accounts for `hard' samples such as selfie pose (cluster 1), back view (cluster 2) and large zoom-in (cluster 3).

\textbf{Pseudo-Label}~
In DFA, each training sample is associated with a pseudo-label that reflects its relationship to the other samples.
Let the ground truth positions of fashion landmarks denote as $l$, where $l_{i}$ specifies the pixel location of landmark $i$.
The pseudo-label,  $f\left(k\right)\in \mathbb{R}^{k\times 1}, k=1\ldots K$, of each sample $i$ is a $K$-dim vector and can be calculated as
\begin{equation}
f\left(k\right) = \exp\left(-\frac{dist\left(i, C\right)}{T}\right),
\end{equation}
where $dist(\cdot)$ is a distance measure. $C_{k}$ is a set of $k$ cluster centers, obtaining by k-means algorithm on the spaces of landmark coordinates (stage-1) or offsets (stage-2 and 3).
$T$ is the temperature parameter to soften pseudo-labels.
We adopt $K = 20$ for all three stages.

Here we explain the pseudo-label in each of the three stages.
Cluster centers $C^{1}_{k}$ in stage-1 are obtained in landmark configuration space $l_{i}$, where $l_{i}$ is the ground truth landmark positions for sample $i$.
Then pseudo-label $f^{1}_{i}$ of sample $i$ in stage-1 can be written as $f^{1}_{i}\left(k\right) = \exp\left(-\frac{\|l_{i} - C^{1}_{k}\|_{2}}{T}\right)$.
We now have a landmark estimation $\hat{l^{1}_{i}}$ from stage-1 for sample $i$.
In stage-2, we first define the landmark offset $\delta l^{2}_{i} = \hat{l^{1}_{i}} - l_{i}$,
which is the correction should be made on stage-1 estimation.
Cluster centers $C^{2}_{k}$ in stage-2 are obtained in landmark offset space $\delta l^{2}_{i}$.
Similarly, pseudo-label $f^{2}_{i}$ of sample $i$ in stage-2 can be written as $f^{2}_{i}\left(k\right) = \exp\left(-\frac{\|\delta l^{2}_{i} - C^{2}_{k}\|_{2}}{T}\right)$.
Since outer product $\otimes$ of two offsets contains the correlations between different fashion landmarks (\eg `left collar' v.s. `left sleeve'),
we further include these contextual information into the pseudo-labels of stage-3 $f^{3}$.
To make the results of outer product comparable, we convert them into vectors by stacking columns, which is denoted as $lin\left(\right)$.
The landmark offset of stage-3 is defined as $\delta l^{3}_{i} = \hat{l^{2}_{i}} - l_{i}$, where $\hat{l^{2}_{i}} = \hat{l^{2}_{i}} + \hat{\delta l^{2}_{i}}$ is the estimation made by stage-2.
Thus, cluster centers $C^{3}_{k}$ in stage-3 are obtained in contextual offset space $\delta_{context} l^{3}_{i} = lin\left(\delta l^{3}_{i} \otimes \delta l^{3}_{i}\right)$.
Similarly, pseudo-label $f^{3}_{i}$ of sample $i$ in stage-3 can be written as $f^{3}_{i}\left(k\right) = \exp\left(-\frac{\|\delta_{context} l^{3}_{i} - C^{3}_{k}\|_{2}}{T}\right)$.
The pseudo-labels used in each stage are summarized in Table \ref{tab:pseudolabels}.

\begin{table}[t]
\scriptsize
\centering
\caption{\footnotesize Summary of pseudo-labels used in each stage. $l_{i}$ is the ground truth landmark positions. $\hat{l^{1}_{i}}$ and $\hat{l^{2}_{i}}$ are the landmark estimations from stage-1 and 2, respectively. $\otimes$ is the outer product and $lin\left(\cdot\right)$ is the linearize operation.}
\label{tab:pseudolabels}
\begin{tabular}{c|c|c|c}
\noalign{\smallskip}
\toprule
~stage~ & ~clustering space~ & ~indication~ & ~dimension~ \\
\midrule\midrule
~1~ & ~$l_{i}$~ & ~landmark configuration~ & 20 \\
\midrule
~2~ & ~$\delta l^{2}_{i} = \hat{l^{1}_{i}} - l_{i}$~ & ~landmark offset~ & 20 \\
\midrule
\multirow{2}{*}{~3~} & ~$\delta_{context} l^{3}_{i} = lin\left(\delta l^{3}_{i} \otimes \delta l^{3}_{i}\right)$~ & \multirow{2}{*}{~contextual offset~} & \multirow{2}{*}{20} \\
 & $\delta l^{3}_{i} = \hat{l^{2}_{i}} - l_{i}$ & & \\
\bottomrule
\end{tabular}
\end{table}

\textbf{Auto-Routing}~
Another important building block of DFA is the auto-routing mechanism.
It is built upon the fact that the estimated pseudo-labels in stage-2 $\hat{f^{2}}$ reflects the error patterns for each sample.
We first associate each cluster center with an average error magnitude $e\left(C^{2}_{k}\right), k = 1\ldots K$.
This can be done by averaging the errors of training samples in each cluster.
Then, we define the error function $G\left(\cdot\right)$ within pseudo-label $\hat{f^{2}_{i}}$ for sample $i$ in stage-2: $G\left(\hat{f^{2}_{i}}\right) = \sum_{k=1}^{K} e\left(C^{2}_{k}\right) \cdot \hat{f^{2}_{i}}\left(k\right)$.
Therefore, the routing function $r_{i}$ for sample $i$ is formulated as
\begin{equation}
r_{i} = \textbf{1}\left(G\left(\hat{f^{2}_{i}}\right) < \epsilon \right),
\end{equation}
where $\textbf{1}\left(\cdot\right)$ is the indicator function and $\epsilon$ is the error threshold for auto-routing.
We set $\epsilon = 0.3$ empirically.
If $r_{i} = 1$ indicates sample $i$ will go through branch $1$ in stage-3, and $r_{i} = 0$ indicates otherwise.

\textbf{Training}~
Each stage of DFA is trained with multiple loss functions, including landmark estimation $L_{positions}$, visibility prediction $L_{visibility}$, and pseudo-label approximation $L_{labels}$.
The overall loss function $L_{overall}$ is
\begin{equation}
L_{overall} = L_{positions}(l, \hat{l}) + \alpha(t)L_{visibility}(v, \hat{v}) + \beta(t)L_{labels}(f, \hat{f}),
\end{equation}
where $\hat{l}$, $\hat{v}$ and $\hat{f}$ are the predicted landmark positions, visibility, and pseudo-labels respectively.
We employ the Euclidean loss for $L_{positions}$ and $L_{labels}$,
and the multinomial logistic loss is adopted for $L_{visibility}$.
$\alpha\left(t\right)$ and $\beta\left(t\right)$ are the balancing weights between them.
All the VGG-16 networks are pre-trained using ImageNet \cite{deng2009imagenet} and the entire DFA cascaded network is fine-tuned by stochastic gradient decent with back-propagation.

The proper scheduling of $\alpha\left(t\right)$ and $\beta\left(t\right)$ is very important for network performance.
If they are too large, it disturbs the training of landmark positions.
If they are too small, the training procedure cannot benefit from these auxiliary information.
Similar to \cite{lee2013pseudo}, we design a piecewise adjustment strategy for $\alpha\left(t\right)$ and $\beta\left(t\right)$ during training process,
\begin{equation}
\alpha\left(t\right) = \left\{
\begin{array}{ll}
\alpha &~~~t < t_{1}, \\
\frac{t-t_{1}}{t_{2}-t_{1}}\alpha &~~~t_{1} \leq t < t_{2}, \\
0 &~~~t_{2} \leq t,
\end{array}
\right.
\end{equation}
where $t_1= 2000$ iterations and $t_2= 4000$ iterations in our implementation. The adjustment for $\beta\left(t\right)$ takes a similar form.

\textbf{Computations}~
For a three stage cascade to predict $8$ fashion landmarks, DeepPose is required to train $17$ VGG-16 models in total, while only three VGG-16 models need to be trained for DFA.
Thus, our proposed approach at least saved $5$ times computational costs.

\section{Experiments}

This section presents evaluation and analytical results of fashion landmark detection, as well as two applications including clothing attribute prediction and clothes retrieval.

\textbf{Experimental Settings}~
For each clothing category, we randomly select $5K$ images for validation and another $5K$ images for test.
The remaining $30K\sim40K$ images are used for training.
We employ two metrics to evaluate fashion landmark detection, normalized error (NE) and the percentage of detected landmarks (PDL) \cite{toshev2014deeppose}.
NE is defined as the $\ell_{2}$ distance between predicted landmarks and ground truth landmarks in the normalized coordinate space (\ie divided by the width/height of the image), while PDL is calculated as the percentage of detected landmarks under certain overlapping criterion.
Typically, smaller values of NE or higher values of PDL indicate better results.

\textbf{Competing Methods}~
Since this work is the first study of fashion landmark detection, it is difficult to find direct comparisons. Nevertheless, to fully demonstrate the effectiveness of DFA, we compare it with two deep models, including DeepPose \cite{toshev2014deeppose} and Image Dependent Pairwise Relations (IDPR) \cite{chen2014articulated}, which achieved best-performing results in human pose estimation. They are two representative methods that explored network cascade and graphical model to handle human pose.
Specifically, DeepPose designed a cascaded deep regression framework on human body parts,
while IDPR combined CNN and structural SVM \cite{tsochantaridis2004support} to model the relations among landmarks.
To have a fair comparison, we replace the backbone networks in DeepPose and IDPR with VGG-16 and carefully train them using the same data and protocol as DFA did.

\subsection{Ablation Study}

\begin{table}[t]
\centering
\caption{\footnotesize Ablation study of DFA on different fashion landmarks. The normalized error (NE) is used here. `abs.' indicates pseudo-labels obtained from absolute landmark positions. `offset' indicates pseudo-labels obtained from local landmark offsets. `c. offset' indicates pseudo-labels obtained from contextual landmark offsets. Intuitively, for the image size of 224$\times$224, the best prediction is achieved in stage-3 using contextual offset as pseudo-labels, whose errors in pixels are 0.048$\times$224=10.752, 10.752, 20.384, 19.936, 15.904, and 16.128 respectively. `T=20' is found empirically in the experiments.}
\label{tab:ablation_landmark}
\scalebox{0.8}{
\begin{tabular}{c|l|ccccccc}
\noalign{\smallskip}
\toprule
~~Stage~~ & ~Component~ & ~L. Collar~ & ~R. Collar~ & ~L. Sleeve~ & ~R. Sleeve~ & ~L. Hem~ & ~R. Hem~ & ~~Avg.~~ \\
\midrule
\midrule
\multirow{4}{*}{1} & ~direct regression~ & .071 & .071 & .134 & .130 & .103 & .102 & .102 \\
  & ~~~$-$ visibility~ & .104 & .102 & .213 & .212 & .141 & .143 & .153 \\\cline{2-9}
  & ~~~+ p.-labels(T=1)~ & .065 & .065 & .113 & .112 & .094 & .095 & .091 \\
  & ~~~+ p.-labels(T=20)~ & \textbf{.057} & \textbf{.058} & \textbf{.106} & \textbf{.104} & \textbf{.088} & \textbf{.089} & \textbf{.084} \\
\midrule
\multirow{3}{*}{2} & ~direct regression~ & .051 & .051 & .098 & .096 & .082 & .080 & .078 \\\cline{2-9}
  & ~~~+ p.-labels(abs.)~ & .050 & .051 & .097 & .096 & .080 & .079 & .076 \\
  & ~~~+ p.-labels(offset)~ & \textbf{.049} & \textbf{.050} & \textbf{.095} & \textbf{.093} & \textbf{.077} & \textbf{.078} & \textbf{.074} \\
\midrule
\multirow{5}{*}{3} & ~direct regression~ & .049 & .049 & .094 & .093 & .074 & .076 & .073 \\\cline{2-9}
  & ~~~+ two-branch~ & \textbf{.048} & .049 & .094 & .093 & .073 & .075 & .072 \\
  & ~~~+ auto-routing~ & \textbf{.048} & .049 & .092 & .091 & .\textbf{071} & .073 & .070 \\\cline{2-9}
  & ~~~+ p.-labels(offset)~ & \textbf{.048} & \textbf{.048} & \textbf{.091} & .090 & \textbf{.071} & .073 & .069 \\
  & ~~~+ p.-labels(c. offset)~ & \textbf{.048} & \textbf{.048} & \textbf{.091} & \textbf{.089} & \textbf{.071} & \textbf{.072} & \textbf{.068} \\
\bottomrule
\end{tabular}
}
\end{table}

We demonstrate the merits of each component in DFA.

\textbf{Effectiveness of Network Cascade}~
Table \ref{tab:ablation_landmark} lists the performance of NE among three stages, where we have two observations.
First, as shown in stage one, training DFA with both landmark position and visibility, denoted as `direct regression', outperforms training DFA with only landmark position, denoted as `- visibility', showing that visibility helps landmark detection because it indicates variations of pose and scale (\eg zoom-in).
Second, cascade networks gradually reduce localization errors on all fashion landmarks from stage one to stage three.
By predicting the corrections over previous stage, DFA decomposes a complex mapping problem into several subspace regression.
For example, Fig. \ref{fig:demo} (e-g) demonstrates the quantitative stage-wise landmark detection results of DFA on different clothing items.
In these cases, stage-1 gives rough predictions with shape constraints, while stage-2 and stage-3 refine results by referring to local and contextual correction patterns.

\begin{table}[t]
\centering
\caption{\footnotesize Ablation study of DFA on different evaluation subsets. NE is used here.}
\label{tab:ablation_subset}
\scalebox{0.8}{
\begin{tabular}{c|l|ccccc}
\noalign{\smallskip}
\toprule
~~Stage~~ & ~Component~ & ~N. Pose~ & ~M. Pose~ & ~L. Pose~ & ~M. Zoom-in~ & ~L. Zoom-in~ \\
\midrule
\midrule
\multirow{2}{*}{1} & ~direct regression~ & .079 & .100 & .111 & .151 & .193 \\
  & ~~~+ p.-label(T=20)~ & \textbf{.064} & \textbf{.077} & \textbf{.085} & \textbf{.120} & \textbf{.156} \\
\midrule
\multirow{2}{*}{2} & ~direct regression~ & .055 & .072 & .081 & .104 & .151 \\
  & ~~~+ p.-label(offset)~ & \textbf{.053} & \textbf{.069} & \textbf{.078} & \textbf{.102} & \textbf{.148} \\
\midrule
\multirow{3}{*}{3} & ~direct regression~ & .052 & .069 & .078 & .096 & .148 \\
  & ~~~+ auto-routing~ & .051 & .067 & .077 & .093 & .146 \\
  & ~~~+ p.-label(c. offset)~ & \textbf{.050} & \textbf{.066} & \textbf{.075} & \textbf{.091} & \textbf{.144} \\
\bottomrule
\end{tabular}
}
\end{table}

\textbf{Effectiveness of Different Pseudo-Labels}~
Within each stage, choices of different pseudo-labels are explored.
We design pseudo-labels representing landmark configurations, local landmark offsets and contextual landmark offsets for three stages respectively.
Table \ref{tab:ablation_subset} shows that pseudo-labels lead to substantial gains beyond direction regression, especially for the first stage.
Next, we further justify the forms of different pseudo-labels adopted for each stage.
In stage-1, we find that using soft label, denoted as `$+p.~labels~(T=20)$', instead of hard label, denoted as `$+p.~labels~(T=1)$', results in better performance,
because soft label is more informative in identifying landmark configuration of samples.
In stage-2, pseudo-label generated from offset landmark positions is superior to that generated from absolute landmark positions since landmark offsets can provide more guidance on the local corrections to be predicted.
In stage-3, including contextual landmark offsets help achieve further gains, due to the fact that landmark corrections to be made are generally correlated.

\textbf{Effectiveness of Auto-Routing}~
Finally, we show that auto-routing is an effective way to tackle data with different correction difficulties.
From Table \ref{tab:ablation_landmark} stage-3, we can see that auto-routing (denoted as `+auto-routing') provides more benefits when compared with averaging the predictions from two branches trained using all data (denoted as `+two-branch').
By further inspecting the stage-wise performance on each evaluation set, which is shown in Table \ref{tab:ablation_subset}, we can observe that auto-routing mechanism improves the performance of medium/large zoom-in subsets, showing that the routing function makes one of the branch in stage-3 focus on difficult samples.

\begin{figure}[t]
  \centering
  \includegraphics[width=0.9\textwidth]{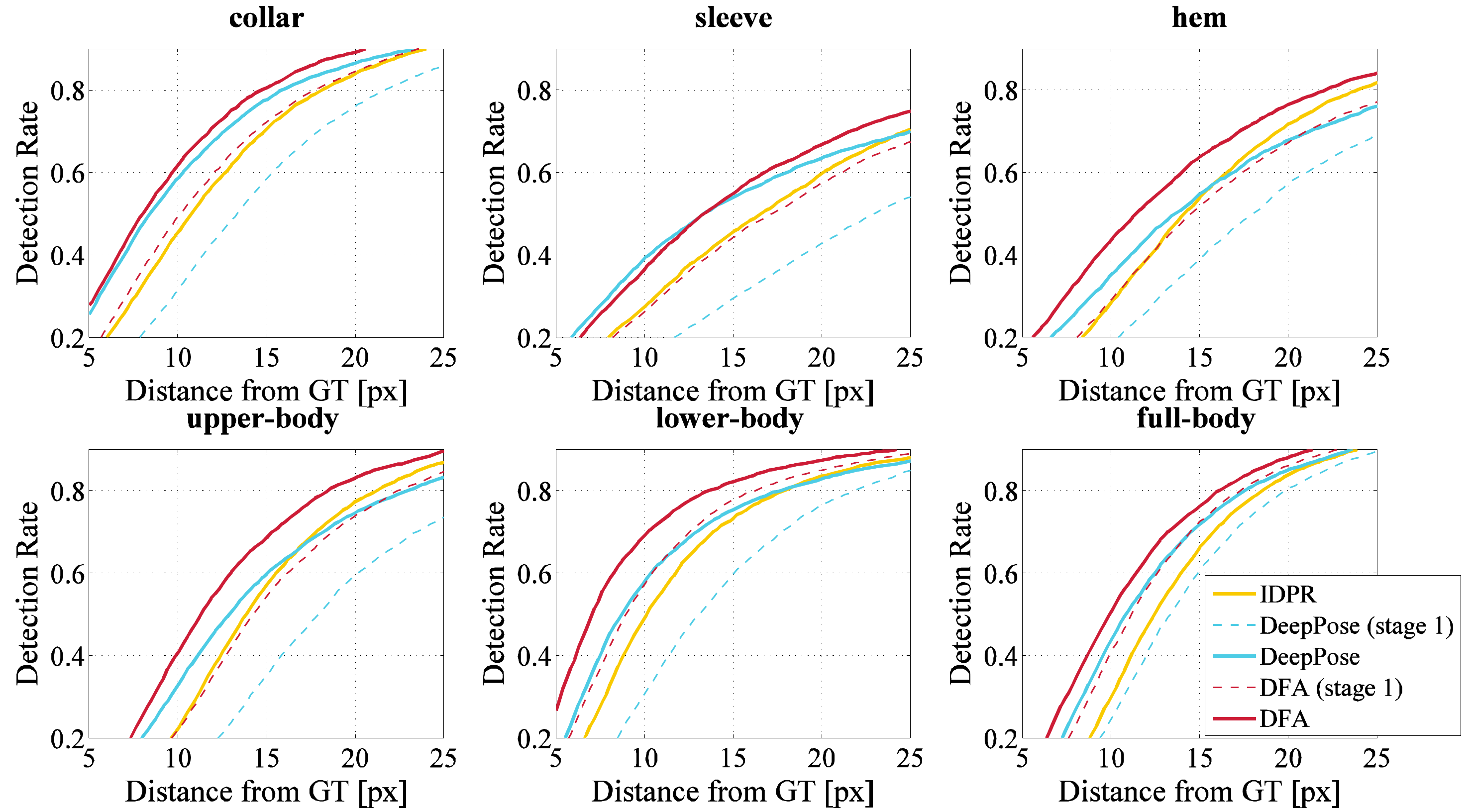}
  \caption{\footnotesize Performance of fashion landmark detection on different fashion landmarks (the first row) and different clothing types (the second row). [px] represents pixels. The percentage of detected landmarks (PDL) is used here.}
  \label{fig:benchmark_landmark}
\end{figure}

\begin{figure}[t]
  \centering
  \includegraphics[width=0.8\textwidth]{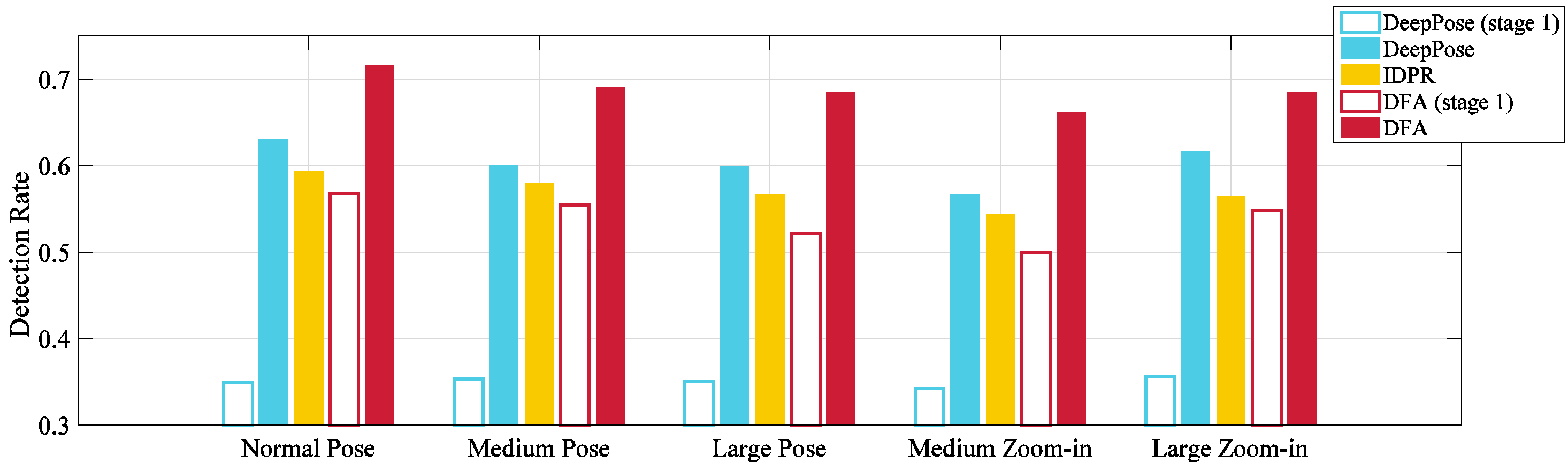}
  \caption{\footnotesize Performance of fashion landmark detection on different evaluation subsets. The distance threshold is fixed at 15 pixels. PDL is used here.}
  \label{fig:benchmark_subset}
\end{figure}

\subsection{Benchmarking}

To illustrate the effectiveness of DFA, we compare it with state-of-the-art human pose estimation methods like DeepPose \cite{toshev2014deeppose} and IDPR \cite{chen2014articulated}.
We also analyze the strengths and weaknesses of each method on fashion landmark detection.

\textbf{Landmark Types}~
Fig. \ref{fig:benchmark_landmark} (the first row) shows the percentage of detection rates on different fashion landmarks, where we have three observations.
First, on landmark `hem', DeepPose performs better when distance threshold is small, while IDPR catches up when the threshold is large,
because DeepPose is a part-based method which can locally refines the results for easy cases.
Second, collars are the easiest landmarks to be detected while sleeves are the hardest.
Third, DFA consistently outperforms both DeepPose and IDPR or shows comparable results on all fashion landmarks, showing that
the pseudo-labels and auto-routing mechanisms enable robust fashion landmark detection.

\textbf{Clothing Types}~
Fig. \ref{fig:benchmark_landmark} (the second row) shows the percentage of detection rates on different clothing types.
Again, DFA outperforms all other methods on all distance thresholds.
We have two additional observations.
First, DFA (stage-1) already achieves comparable results on full-body and lower-body clothes when compared with IDPR and DeepPose (stage-3).
Second, upper-body clothes pose most challenges on fashion landmark detection.
It is partially due to the various clothing sub-categories contained.

\textbf{Difficulty Levels}~
Fig.\ref{fig:benchmark_subset} shows the percentage of detection rates on different evaluation subsets, with the distance threshold fixed at 15 pixels.
Two observations are made here.
First, fashion landmark detection is a challenging task.
Even the detection rate for normal pose set is just above $70\%$.
More powerful model needs to be developed.
Second, DFA has the most advantages on medium pose/zoom-in subsets.
Pseudo-labels provide effective shape constraints for hard cases.
Please also note that DFA (stage-3) requires much less computational costs than DeepPose (stage-3).

For a $300 \times 300$ image, DFA takes around $100ms$ to detect full sets of fashion landmarks on a single GTX Titan X GPU.
In contrast, DeepPose needs nearly $650ms$ in the same setting.
Our framework has large potential in real-life applications.
Visual results of fashion landmark detection by different methods are given in Fig.\ref{fig:demo}.

\begin{table}[t]
\scriptsize
\centering
\caption{\footnotesize Comparison of strict PCP results on the LSP dataset. DFA shows good generalization ability to human pose estimation.}
\label{tab:generalization}
\begin{tabular}{l|ccccccc}
\noalign{\smallskip}
\toprule
~Method~ & ~Torso~ & ~Head~ & ~U.arms~ & ~L.arms~ & ~U.legs~ & ~L.legs~ & ~Mean~ \\
\midrule
\midrule
~Fu \etal \cite{fu2015beyond} & 77.7 & 85.4 & 75.0 & 71.9 & 62.1 & 48.8 & 67.7 \\
~Ouyang \etal \cite{ouyang2014multi} & 85.8 & 83.1 & 63.3 & 46.6 & 76.5 & 72.2 & 68.6 \\
~Pose Machines \cite{ramakrishna2014pose} & 93.1 & 83.6 & 76.8 & 68.1 & 42.2 & 85.4 & 72.0 \\
~DeepPose \cite{toshev2014deeppose} & - & - & 56 & 38 & 77 & 71 & - \\
~DFA & 90.8 & \textbf{87.2} & 70.4 & 56.2 & \textbf{80.6} & 75.8 & 74.4 \\
\midrule
~IDPR \cite{chen2014articulated} & 92.7 & 87.8 & 69.2 & 55.4 & 82.9 & 77.0 & 75.0 \\
~~~+ p.-label~ & 93.5 & 88.5 & 72.3 & 59.0 & 83.9 & 78.7 & 77.0 \\
~~~+ auto-routing~ & \textbf{94.1} & \textbf{88.9} & \textbf{74.3} & \textbf{61.5} & \textbf{85.1} & \textbf{80.4} & \textbf{78.6} \\
\bottomrule
\end{tabular}
\end{table}

\subsection{Generalization of DFA}

To test the generalization ability of the proposed framework, we further apply DFA on a related task, \ie human pose estimation, as reported in Table \ref{tab:generalization}.
In the following, DFA is trained and evaluated on LSP dataset \cite{johnson2010clustered} as \cite{chen2014articulated} did.

First, we compare DFA system with other state-of-the-art methods on pose estimation task.
Without much adaptation, DFA achieves $74.4$ mean strict PCP results, with $87$, $91$, $70$, $56$, $81$, $76$ for `torso', `head', `u.arms', `l.arms', `u.legs' and `l.legs' respectively. It shows comparable results to \cite{chen2014articulated} and outperforms several recent works \cite{fu2015beyond,ouyang2014multi,ramakrishna2014pose,toshev2014deeppose}, showing that
DFA is a general approach for structural prediction problem besides fashion landmark detection.

Then, we show that pseudo-label and auto-routing scheme of DFA can be generalized to improve performance of pose estimation methods, such as IDPR \cite{chen2014articulated}.
\cite{chen2014articulated} trained DCNN and achieved $75$ mean strict PCP. We add pseudo-labels to this DCNN and include auto-routing in cascading predictions. Training and evaluation of graphical model are kept unchanged. Pseudo-labels leverage the result to $77$ mean strict PCP and auto-routing leads to another $1.6$ point gain.
It demonstrates that pseudo-labels and auto-routing are effective and complementary techniques to current methods.

\begin{figure}[t]
  \centering
  \includegraphics[width=0.9\textwidth]{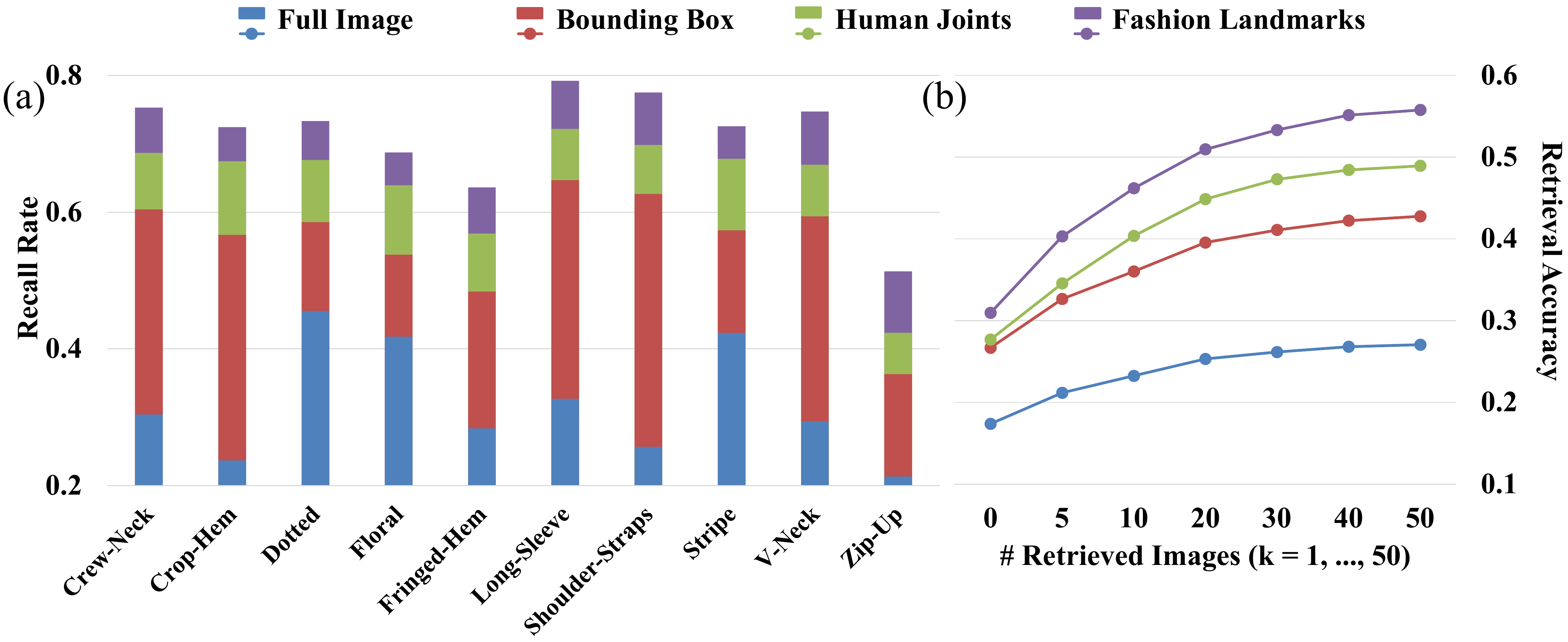}
  \caption{\footnotesize Experimental results on clothing attribute prediction and clothes retrieval using features extracted from `full image', `bbox', `human joints', and `fashion landmarks': (a) the top-$5$ recall rates of clothing attributes and (b) the top-$k$ clothes retrieval accuracy.}
  \label{fig:application}
\end{figure}

\subsection{Applications}

Finally, we show that fashion landmarks can facilitate clothing attribute prediction and clothes retrieval.
We employ a subset of DeepFashion dataset \cite{liu2016deepfashion}, which contains $10K$ images, $50$ clothing attributes and corresponding image pairs (\ie the images containing the same clothing item).
We compare fashion landmarks with different localization schemes, including the full image, the bounding box (bbox) of clothing item, and the human-body joints, where
fashion landmarks are detected by DFA, human joints are obtained by the executable code of \cite{chen2014articulated}, and bounding boxes are manually annotated.
For both tasks of attribute recognition and clothes retrieval, we use off-the-shelf CNN features as described in \cite{simonyan2014very}.

\textbf{Attribute Prediction}~
We train a multi-layer perceptron (MLP) using the extracted CNN features as input to predict all $50$ attributes.
Following \cite{gong2013deep}, we employ the top-$k$ recall rate as measuring criteria, which is obtained by ranking the classification scores and determine how many ground truth attributes have been found in the top-$k$ predicted attributes.
Overall, the average top-$5$ recall rates on 50 attributes of `full image', `bbox', `human joints', and `fashion landmarks' are $27\%$, $53\%$, $65\%$ and $73\%$, respectively, showing that fashion landmarks are the most effective representation for attribute prediction of fashion items.
Fig.\ref{fig:application} (a) shows the top-$5$ recall rates of ten representative attributes, \eg `stripe', `long-sleeve', and `V-neck'.
We observe that fashion landmark outperforms all the other localization schemes in all the attributes, especially for part-based attributes, such as `zip-up' and `shoulder-straps'.

\textbf{Clothes Retrieval}~
We adopt the $\ell_{2}$ distance between the extracted CNN features for clothes retrieval.
The top-$k$ retrieval accuracy is used to measure the performance, such that given a query image, correct retrieval is considered as the exact clothing item has been found in the top-$k$ retrieved gallery images.
As shown in Fig.\ref{fig:application} (b),
the top-$20$ retrieval accuracies for `full image', `bbox', `human joints', and `fashion landmarks' are $25\%$, $40\%$, $45\%$, and $51\%$ respectively.
When $k=1$ and $k=5$, features extracted around fashion landmarks still perform better than the other alternatives, demonstrating that fashion landmarks provide more discriminative information beyond traditional paradigms.

\begin{figure}[t]
  \centering
  \includegraphics[width=1.0\textwidth]{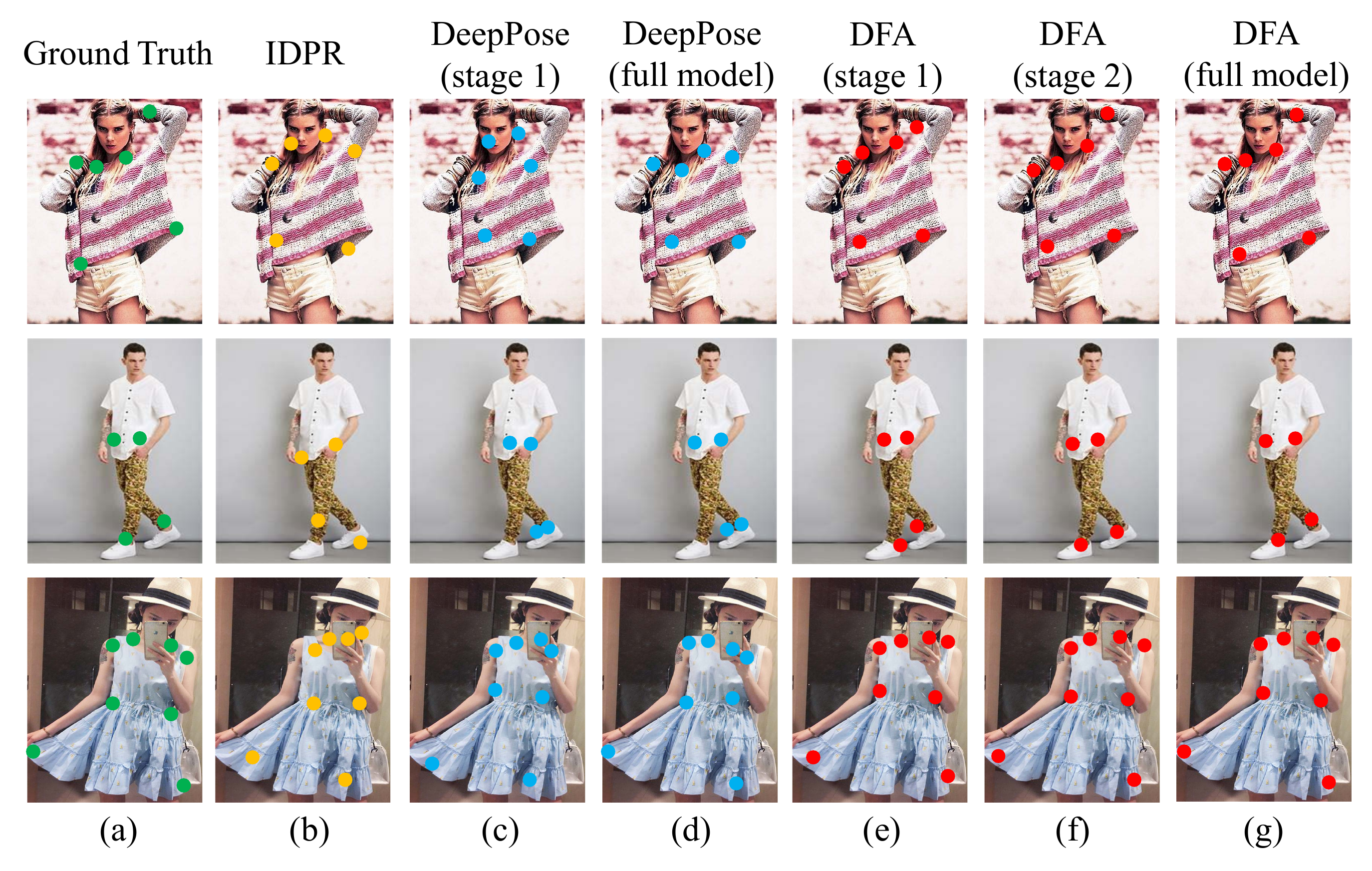}
  \caption{\footnotesize Visual results of fashion landmark detection by different methods: (a) Ground Truth, (b) IDPR, (c) DeepPose (stage 1), (d) DeepPose (full model), (e) DFA (stage 1), (f) DFA (stage 2) and (g) DFA (full model).}
  \label{fig:demo}
\end{figure}

\section{Conclusions}

This paper introduced fashion landmark detection, which is an important step towards robust fashion recognition.
To benchmark fashion landmark detection, we introduced a large-scale fashion landmark dataset (FLD).
With FLD, we proposed a deep fashion alignment network (DFA) for robust fashion landmark detection, which leverages pseudo-labels and auto-routing mechanism to reduce the large variations presented in fashion images.
Extensive experiments showed the effectiveness of different components as well as the generalization ability of DFA.
To demonstrate the usefulness of fashion landmark, we evaluated on two fashion applications, clothing attribute prediction and clothes retrieval.
Experiments revealed that fashion landmark is a more discriminative representation than clothes bounding boxes and human joints for fashion-related tasks, which we hope could facilitate future research.

%\section*{\footnotesize{Acknowledgements}}
\scriptsize{\textbf{Acknowledgements} This work is partially supported by SenseTime Group Limited, the Hong Kong Innovation and Technology Support Programme, the General Research Fund sponsored by the Research Grants Council of the Kong Kong SAR (CUHK 416312),  the External Cooperation Program of BIC, Chinese Academy of Sciences (No.172644KYSB20150019),
the Science and Technology Planning Project of Guangdong Province (2015B010129013, 2014B050505017), and the National Natural Science Foundation of China (61503366, 61472410; Corresponding author is Ping Luo).}

\clearpage

\bibliographystyle{splncs}
\bibliography{egbib}

\clearpage

\begin{appendix}

\normalsize

\textbf{\Large Appendices}

\section{More Comparisons between DFA and DeepPose}

\begin{table}
\begin{center}
\begin{tabular}{c|c|c|c|c|c|c|c}
\hline
\multicolumn{2}{c|}{} & \multicolumn{2}{|c|}{~stage-1~} & \multicolumn{2}{|c|}{~stage-2~} & \multicolumn{2}{|c}{~stage-3~} \\
\hline
\multicolumn{2}{c|}{} & ~DFA~ & ~DeepPose~ & ~DFA~ & ~DeepPose~ & ~DFA~ & ~DeepPose~ \\
\hline\hline
\multirow{4}{*}{~input~} & ~image~ & \checkmark & \checkmark & \checkmark & & \checkmark & \\
 & ~local patches~ & & & & \checkmark & & \checkmark \\
 & ~estimated pos.~ & & & \checkmark & & \checkmark & \\
 & ~pseudo-labels~ & & & & & \checkmark & \\
\hline
\multirow{2}{*}{~supervision~} & ~pos.+vis.~ & \checkmark & \checkmark & \checkmark & \checkmark & \checkmark & \checkmark \\
& ~offsets~ &  &  & \checkmark & \checkmark & \checkmark & \checkmark \\
 & ~pseudo-labels~ & \checkmark & & \checkmark & & \checkmark & \\
\hline
~\# CNNs~ & & 1 & 1 & 1 & 8 & 2 & 8 \\
\hline
\end{tabular}
\end{center}
\caption{Comparisons between the pipelines of DFA and DeepPose. `pos.' and `vis.' represent landmark positions and visibility.}
\label{tab:pipeline}
\end{table}

More comparisons between DFA and DeepPose are given in Table \ref{tab:pipeline}.

\begin{figure}[t]
  \centering
  \includegraphics[width=0.9\textwidth]{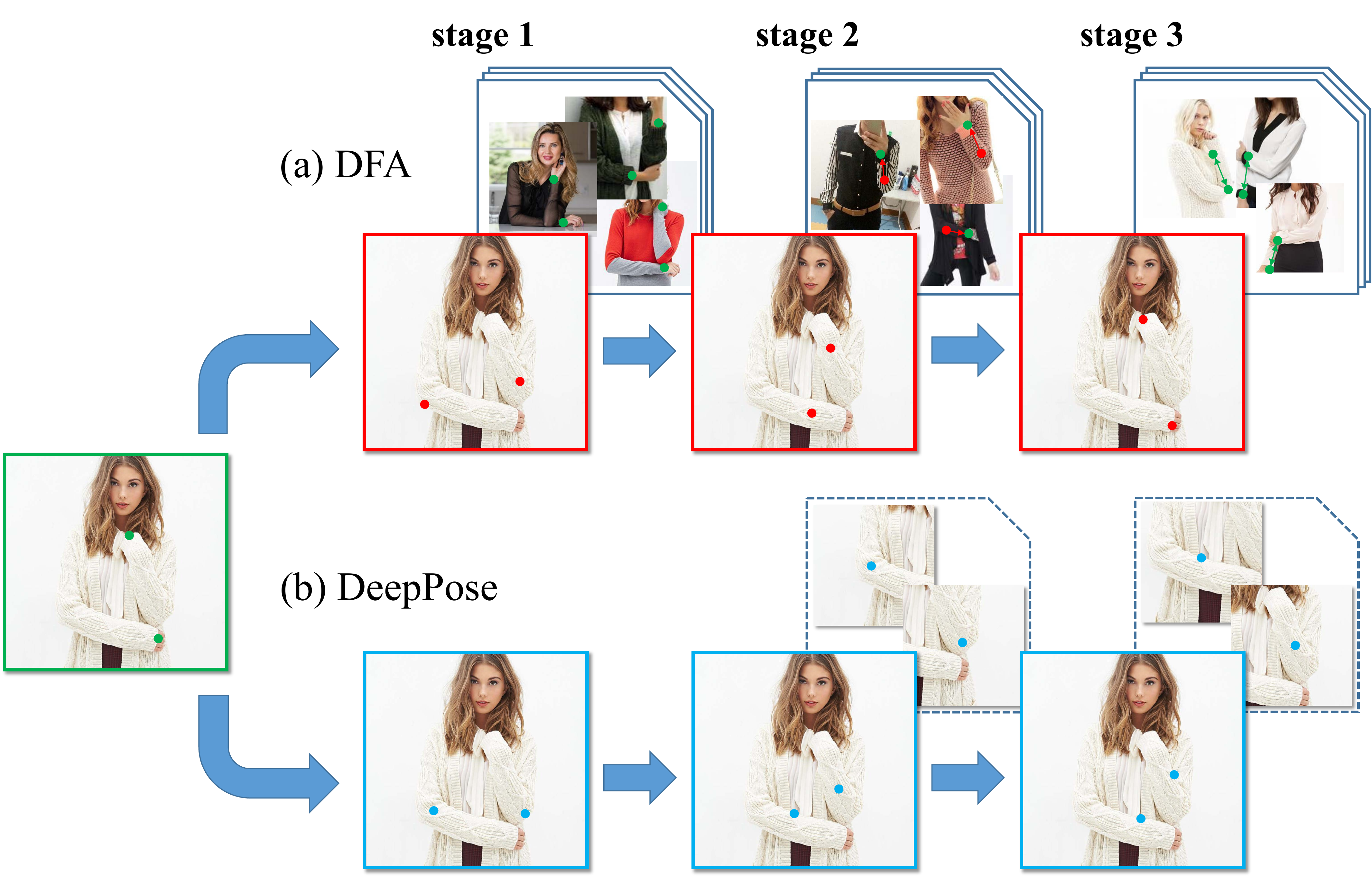}
  \caption{Comparison of DFA and DeepPose: (a) the fashion landmark detection results of DFA in each stage, along with the images with similar pseudo-labels to the input image; (b) the landmark detection results of DeepPose in each stage, along with input patches for stage-2 and 3. Only `left sleeve' and `right sleeve' are illustrated.}
  \label{fig:pipeline_comparison}
\end{figure}

For all the three stages, DFA employs the images as input while DeepPose uses the local image patches (\ie the regions predicted by the previous stage) except stage-1. The last two stages of DFA also employ the estimated landmarks' positions as additional input signal. As shown in the Fig.1 (b) and (c) in the paper, the local patches of fashion images have large variations of scale and appearance. They loss much of the discriminative information, such that using the local patches as input may harm the accuracy. For example, as demonstrated in Fig.\ref{fig:pipeline_comparison} (b) below, where the `right sleeve' takes place in an unusual pose.
Since the initial estimation in stage-1 is far from ground truth, it is hard for DeepPose (stage-2 and 3) to correct only based on the local observations.

DFA employs the full images as input other than local patches, without losing the context information. Furthermore, the estimated landmark positions of the previous stage and the pseudo labels implicitly capture global shape structure and local shape constraints respectively, which facilitate to improve performance.
For example, as shown in Fig.\ref{fig:pipeline_comparison} (a), we see that pseudo-labels in stage-1 mainly reflect holistic pose like hold arms.
Pseudo-labels in stage-2 instead suggest possible ways to correct initial estimations, \eg stretching the estimations of `sleeve' further to the end.
Moreover, pseudo-labels in stage-3 incorporate pairwise relationships between fashion landmarks, \eg the relative position from `left sleeve' to `right sleeve' here.
Combining all three guidance leads to the effectiveness and robustness of DFA.

The carefully designed architecture of DFA reduces the number of VGGs in the network cascade, increasing efficiency of both the training and testing procedures.

\section{Implementation Details}

In this section, we introduce the implementation details of Deep Fashion Alignment (DFA) and DeepPose \cite{toshev2014deeppose}.

\subsection{Pre-processing}

Both Deep Fashion Alignment (DFA) and DeepPose adopt the cascading framework.
In each stage, VGG16 \cite{simonyan2014very} is used as the regression model $l_{i} = \Psi\left(I; \theta\right)$, where $I$ is the input image and $\theta$ is the network parameters.
We replace the original $1000$-way classification layer with one $16$-way landmark position regression layer and eight $3$-way landmark visibility classification layers.
Clothes bounding box is taken as the initial input
while both landmark positions and visibility act as the supervision signals for training.
Please note that the truncated landmarks doesn't induce any error for the landmark position regression.

We also normalize the landmark positions following \cite{toshev2014deeppose}.
Assume a clothes bounding box is defined by its center $\left(x_{c}, y_{c}\right)$ as well as width $b_{w}$ and height $b_{h}$: $b = \left(x_{c}, y_{c}, b_{w}, b_{h}\right)$.
Then our final landmark position $l_{i} = \left(x_{i}, y_{i}\right)$ for landmark $i$ can be obtained by normalizing the absolute landmark coordinates $l_{i}^{orig} = \left(x_{i}^{orig}, y_{i}^{orig}\right)$:
\begin{equation}
\left(x_{i}, y_{i}\right) = \mathcal{N}\left(l_{i}; b\right) = \left(
\begin{array}{cc}
1/b_{w} & 0 \\
0 & 1/b_{h}
\end{array}
\right)
\left[\left(x_{i}^{orig}, y_{i}^{orig}\right) - \left(x_{c}, y_{c}\right)\right]
\end{equation}
where $\mathcal{N}\left(\cdot\right)$ is the normalization function.
After inference, the estimated landmarks in absolute coordinates $\hat{l}_{i}^{orig}$ can be readily read as:
\begin{equation}
\hat{l}_{i}^{orig} = \mathcal{N}^{-1}\left(\Psi\left(I; \theta\right); b\right)
\end{equation}
where $\mathcal{N}^{-1}\left(\cdot\right)$ is the inverse operation of normalization function $\mathcal{N}\left(\cdot\right)$.
Next, we introduce the detailed implementation pipeline for DPA and DeepPose respectively.

\subsection{Input Preparation for DeepPose}

DeepPose (stage-1) takes the clothing bounding box as input and regresses all $N$ fashion landmarks within its receptive field.
These estimated landmark positions are used for the input preparation in subsequent stages.
Then, there will be $N$ models in DeepPose (stage-2).
For each model in stage-2, the input is part bounding box cropped around the estimated fashion landmark (\eg `left sleeve').
The task is to regress the offset for the underlying fashion landmark.
The size of the bounding box is set to be $120\times120$, such that more contextual information can be included.
Similar procedure applies for stage-3 for further refinement.

\section{More Results}

\begin{figure}[t]
  \centering
  \includegraphics[width=1.0\textwidth]{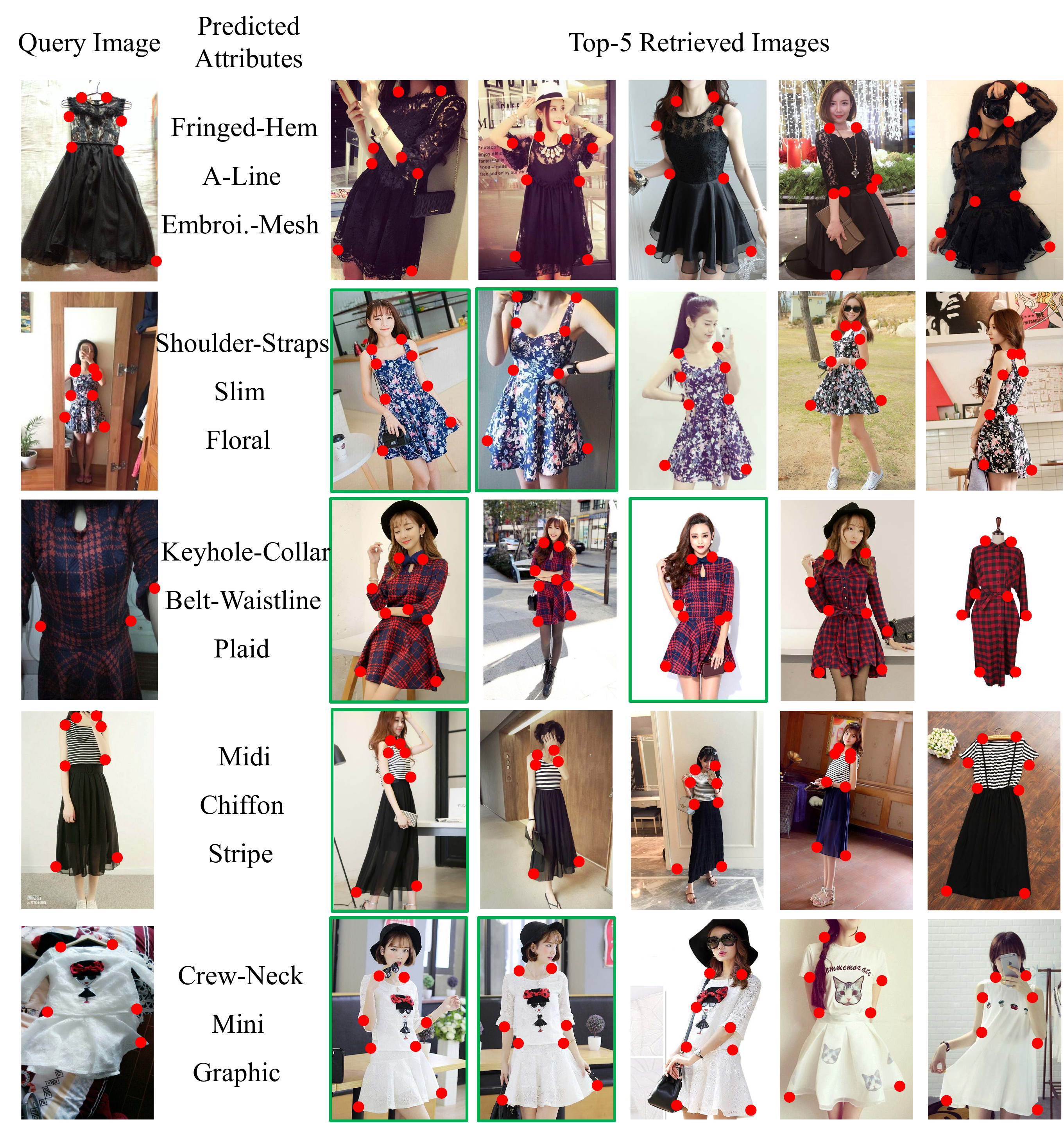}
  \caption{Additional attribute prediction and clothes retrieval results by DFA, along with their predicted landmarks. Correct matches are marked in green.}
  \label{fig:demo_retrieval}
\end{figure}

\begin{figure}[t]
  \centering
  \includegraphics[width=1.0\textwidth]{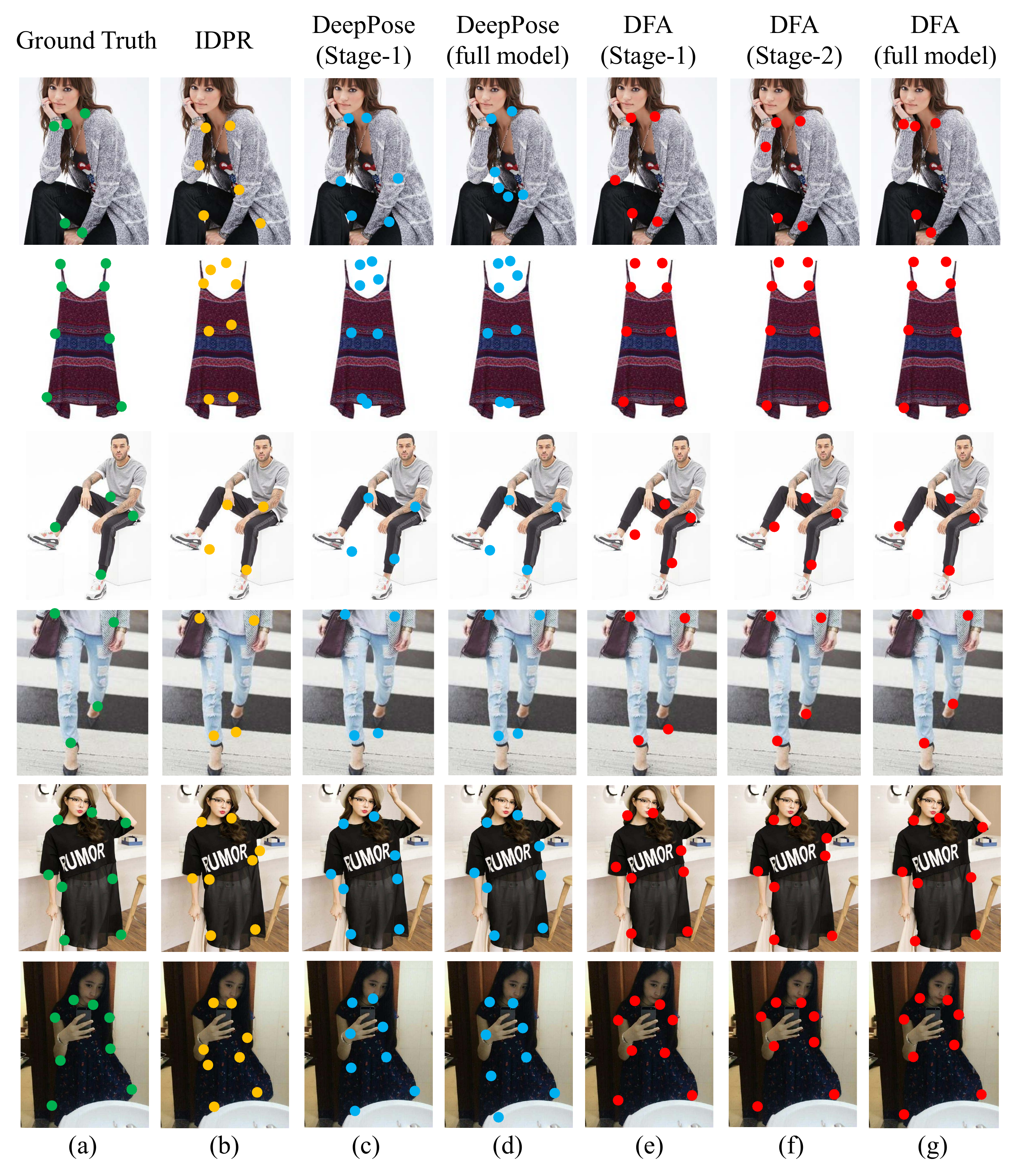}
  \caption{Additional visual results of fashion landmark detection by different methods: (a) Ground Truth, (b) IDPR, (c) DeepPose (stage-1), (d) DeepPose (full model), (e) DFA (stage-1), (f) DFA (stage-2) and (g) DFA (full model).}
  \label{fig:demo_landmarks}
\end{figure}

Fig.\ref{fig:demo_retrieval} and Fig.\ref{fig:demo_landmarks} demonstrate more visual results on attribute prediction/clothes retrieval and fashion landmark detection, respectively.
DFA is capable of handling complex variations under different scenarios.

\end{appendix}

\end{document}